\documentclass[12pt]{article}

\makeatletter
\newcommand\ackname{Acknowledgements}
\if@titlepage
  \newenvironment{acknowledgements}{%
      \titlepage
      \null\vfil
      \@beginparpenalty\@lowpenalty
      \begin{center}%
        \bfseries \ackname
        \@endparpenalty\@M
      \end{center}}%
     {\par\vfil\null\endtitlepage}
\else
  \newenvironment{acknowledgements}{%
      \if@twocolumn
        \section*{\abstractname}%
      \else
        \small
        \begin{center}%
          {\bfseries \ackname\vspace{-.5em}\vspace{\z@}}%
        \end{center}%
        \quotation
      \fi}
      {\if@twocolumn\else\endquotation\fi}
\fi
\makeatother

\begin{document}

\title{Parsing using a grammar of word association vectors}

\author{Robert John Freeman\\ rob@chaoticlanguage.com}

\maketitle

\begin{abstract}
This paper was was first drafted in 2001 as a formalization of the system described in U.S. patent U.S. 7,392,174. It describes a system for implementing a parser based on a kind of cross-product over vectors of contextually similar words. It is being published now in response to nascent interest in vector combination models of syntax and semantics. The method used aggressive substitution of contextually similar words and word groups to enable product vectors to stay in the same space as their operands and make entire sentences comparable syntactically, and potentially semantically. The vectors generated had sufficient representational strength to generate parse trees at least comparable with contemporary symbolic parsers.
\end{abstract}

\section{Motivation}

\subsection{Basic intuition}

The basic intuition underlying our formalisation is that a word can be
characterised by enumerating words that occur in similar
contexts, e.g. Menno van Zaanen\cite{zaanen-bootstrapping} attributes to Harris\cite{harris-msl} the "notion of interchangeability":

\begin{quote}
constituents of the same type can be replaced by each other
\end{quote}

There are variations on this (c.f. if constituents are not of the same type they involve a change of meaning and cannot be replaced), but it involves a comparison, and we will refer to it in general as the "contrastive method".

This intuition is implicitly present in almost all linguistic
classifications. E.g. the definition of a part of speech such as {\em
determiners} is based on the observation that there is a group of
words that exhibit similar behaviour in a range of contexts:

\begin{itemize}
\item preceeding \{"cat", "dog", "car", "wardrobe"\dots\}
\item preceeding \{"big", "small", "blue", "bright"\dots\}
\item following \{"take", "put", "see"\dots\}
\item \dots
\end{itemize}
Of course, in the traditional way of classifying language units such
observations are followed by very strong generalisations. But is this reasonable?

\subsection{Idiosyncratic usage}

A strong theme in linguistics for many years has been the idiosyncrasy of language use defying classification. For an example which appears to date to Halliday\cite{powerfultea}, "strong" and "powerful" can be used interchangeably in many contexts, but not in the context of "tea", viz. "strong tea"/"*powerful tea". Pawley and Syder\cite{two-puzzles} present long lists of examples which defy codification in rules under the title of "native-like selection", which they characterize as a "puzzle" for linguistics.

In fact the Applied Linguistics literature is full of observations that language appears to have no rules without exceptions, and indeed in some senses to consist of exceptions only. Nattinger\cite{nattingerLPG}, Weinert\cite{weinertRFL}, and Lewis\cite{lewisLA} provide a sampling, reaching an extreme with Lewis's "Lexical Approach".

This is addressed less often in a machine learning context, but even here Dagan, Marcus, Markovitch note, Dagan\cite{dagan95contextual}:

\begin{quote}
It has been traditionally assumed that ... information about words should be generalized using word classes... However, it was never clearly shown that unrestricted
language is indeed structured in accordance with this assumption.
\end{quote}

\subsection{The phoneme}

But we can go further than the lack of proof noted by Dagan et al. It was shown not to be so. In fact, you could characterize this is as the central dilemma of linguistics for the last 50 years. It is behind the major divisions which have characterized the subject during that time. Put more strongly, what we find when we try to abstract language structure from observation is that it contradicts itself. And this was shown to be so a long time ago for the first category, and great success of the contrastive method, the phoneme.

The crisis was catalysed by the polarizing figure of 20th century linguistic theory, Noam Chomsky. Chomsky brought a fresh perspective to many issues in linguistics, but while it has been largely forgotten it is fair to say his attack on the phoneme was one of the most important at the time. E.g. Hill\cite{archHilldarling}:

\begin{quote}
I could stay with the Transformationalists pretty well, until they attacked my darling, the phoneme.
\end{quote}

Contrastive methods along the lines attributed to Harris above were the standard method of linguistic analysis. The phoneme was the major success of the method. Chomsky pointed out that these contrastive methods led to contradictory results, and linguistics shattered. As Newmeyer\cite{newmeyerHistPerspective} says while discussing Chomsky's Logical Basis of Linguistic Theory\cite{chomskyLBLT}

\begin{quote}
Part of the discussion of phonology in 'LBLT' is directed towards showing that the conditions that were supposed to define a phonemic representation (including complementary distribution, locally determined biuniqueness, linearity, etc.) were inconsistent or incoherent in some cases and led to (or at least allowed) absurd analyses in others.
\end{quote}

So abstracting language categories from observation led to "inconsistent or incoherent ... analyses". Or equivalently that elements of language combine non-linearly. Lamb\cite{lamblinearity} says of Chomsky:

\begin{quote}
He correctly pointed out that the usual solution incorporates a loss of generality, but he misdiagnosed the problem. The problem was the criterion of linearity.
\end{quote}

See also Lamb 1966\cite{lambPTP}, Lamb 1967\cite{lambAA}, Lamb 1976\cite{lambvanderslice}, Lamb 1998\cite{lambdrummer}.

\subsection{The (very) short history of machine learning in theoretical linguistics}
This was shown to be the case for the phoneme, but it was not seriously contested for other categories. The contrastive method never got beyond the phoneme. Instead linguistics moved beyond abstraction of categories by contrastive analysis, fragmenting into schools which on the one hand continued to respect contrastive analysis but rejected structure as a meaningful parameter, like Functionalism, or on the other hand continued to respect language structure, but rejected the idea it could be learned from observations, Generativism. Generativism hypothesized that the contradictions observed when you tried to abstract categories from observations were not relevant to the system of language at all. It was still assumed structure was relevant, and objective, but because learnable categories had been observed to contradict (among other reasons), structure was assumed to be innate. 

Unfortunately, having split the field, this key issue of contradictions in learnable categories was largely forgotten and unavailable to those outside linguistics seeking insights. So when rapid developments in computing made large amounts of data available, non-linguists started trying to learn from it. Non-linguists in the main, because to reiterate, theoretical linguistics had already split into schools which assumed language structure was unlearnable, or irrelevant, making the issue disappear. It is ironic that for this very reason the non-linguists perhaps felt linguistics was irrelevant and ignored it, in its turn, because it did not address the issue which interested them viz. learning categories from observable data.

\subsection{Contradiction in language categories}
This then is our motivation for seeking to model grammar as a vector product of word similarity vectors, most forcefully that abstractions from observations were shown to lead to contradictions, or in the case of Lamb "non-linearity". That this is so is actually a rare point of agreement in theoretical linguistics, it is only interpretations which have varied: Universal Grammar, Functionalism, Cognitivism.

Unfortunately for a rare point of agreement in linguistics, it is widely ignored. Notably it appears to be ignored in the field of machine learning. It might be seen to render the entire field of grammatical induction moot.

Chomsky concluded that if a coherent language structure could not be learned from observations, then it must be innate. Functionalists (and various schools of Cognitivism) concluded it was irrelevant. But unlearnable need not mean innate or irrelevant. It may simply mean generalizations can only be ad-hoc. We have a better understanding of complexity in formal systems now. Unlearnable, random (in a specific technical sense e.g. Kolmogorov\cite{kolmogorovTRN}, Chaitin\cite{chaitinUnknowable}), or even chaotic, systems are starting to seem more like the norm. As far as the author is aware these kinds of complexity issues were first suggested, at least in the context of computational solutions, in an earlier discussion of the motivations surrounding this parser in Freeman 2000\cite{freemanNAACL}. But it is possible to find similar ideas in linguistics, notably Hopper's Emergent Grammar 1987\cite{hopperEG}:

\begin{quote}
The notion of emergence is a pregnant one. It is not intended to be a
standard sense of origins or genealogy, not a historical question of 'how'
the grammar came to be the way it 'is', but instead it takes the adjective
emergent seriously as a continual movement towards structure, a postponement
or 'deferral' of structure, a view of structure as always provisional, always
negotiable, and in fact as epiphenomenal, that is at least as much an effect
as a cause.
\end{quote}

Possibly Lamb's "non-linearity" is the same.

As suggested in that 2000 paper, this ad-hoc character should by no means be regarded as a bad thing. Rather it hints at a far greater richness of structure available to language.

\section{Formalization}

According to the method of this paper then, in order to prevent relevant information from being lost in the task of modelling natural language, we need a
concept of overlapping classes that can be constructed, ad-hoc, on the fly as
opposed to having to be pre-defined.

\subsection{Similarity}

Thus we stop at the primary
observation that ``word $w$ is similar to the words $w_1,...,w_k$
because it appears in similar contexts'' and treat $w, w_1,...,w_k$ as
class label for $w$.

To put this more formally, we need to define two things: what is
actually a context and what {\em similar contexts} means. A simple
initial approximation to the former question is to define the context
of a word (or word group) $w$ as the set of all word pairs $w', w''$
such that there is at least one occurence of the triplet $w',w,w''$ in
the available corpus.
\[
Con(w) := \{(w',w''):\exists i: X_i=w', X_{i+1}=w, X_{i+2}=w''\}
\]
Now the similarity of two words $w$ and $w'$ can be measured by the
percentage of common contexts relative to all contexts they appear in. Dekang Lin provides a principled measure\cite{linSim}:
\[
sim(w,w') := \frac{2 \times I(Con(w) \cap Con(w'))}{I(Con(w)) + I(Con(w'))}
\]
Where $I$ is the information.

With the two definitions formulated above, the class a word belongs to
can be defined as a set of similarity values between words as below. For convenience we limit the set to include only words similar beyond a threshold $C$:
\[
class(w) := \{w': sim(w,w') > C \}
\]
e.g. A segment from the entry from "somewhat" estimated for one of our corpora is:

\small
\begin{verbatim}
somewhat {somewhat,1 significantly,0.047 slightly,0.043
considerably,0.036 substantially,0.025 far,0.025 a&lot,0.024
more&and,0.024 one&or,0.024 much,0.022 becoming,0.021
likely&to&be,0.021 a&little,0.02 considered,0.018 nothing,0.017
rather,0.017 so&much,0.016 relatively,0.016 something,0.015
certainly,0.014 quite,0.014 generally,0.014 getting,0.014 still,0.012
therefore,0.011 done,0.011 often,0.011 of&course,0.01 even,0.0097
very,0.0083 always,0.0082 a&very,0.008 once,0.0073
further,0.0072 little,0.0071 less,0.007 really,0.0066...}

\end{verbatim}
\normalsize

Where the vector is made up of a set of (word, score) pairs, and the
score is the value of the chosen similarity function between the pair "somewhat" and
the other words in the vocabulary.

For practical reasons it is more convenient to
represent the class of word $w$ as a vector of real numbers, each
representing the similarity of a vocabulary word to $w$. So if
$w_1,...,w_N$ is the vocabulary of our language, then the class
of $w$ is defined by the following function $\phi$:

\[
\phi(w) := (\phi_1(w),...,\phi_N(w))
\]

\subsection{Application to trees}

We now extend the formalization to estimate representations for potentially unobserved sequences of words. We say ``potentially'' unobserved, because the sequences may in fact be observed, however we seek to extend the representation in the absence of observation. Essentially we seek to represent syntax, specifying which new combinations of words are acceptable to the language.

We will call these potentially unobserved sequences $t$ for “trees”, because the formula is non-associative. Different orders give different products, so the product has structure, like a tree. This can be seen as an advantage of the method, it predicts phrase structure.

In the vector formulation, we have a vocabulary V:
$w_1,...,w_N$ of words
and word sequences. Let the function $\delta(i,j) \rightarrow \{0,...,N\}$ map a
pair of elements in V to another entry in V. Simply put, let it map a pair into an entry for the pair together, if they are observed to occur together
($\delta(i,j)=0$ means that no such entry exists, i.e. i and j are not observed to occur together.)

Formally, the extension of the word class function to binary trees is defined recursively, if $t$ is a word, then $\phi_i(t)$ reduces to $sim(t,w_i)$, if t is a binary tree consisting of subtrees $t_1$ and $t_2$, then we define the components of a new vector representing the combination of trees to be:

\[
\phi_k(t):= \sum_{i,j} MI(w_i,w_j) \cdot \phi_i(t_1) \cdot \phi_j(t_2)
\cdot \phi_k(\delta(i,j))
\]

In words: if a pair of words between the components of the operand trees are observed to occur together ($\delta(i,j) \neq 0$), the components for their observed combination ($\phi_k(\delta(i,j))$) contribute to a new component for the combination of the operand trees ($\phi_k(t)$.) Note: $MI$ is the common "mutual information" between two words occurring in sequence, which was used in this case to scale the significance of an observed pairing. We in no way regard this choice as exhaustive or exclusive. In much the same way as the similarity measure described in section 2.1 above, many scalar association measures might be considered.

Essentially we seek to generate product vectors by an aggressive substitution of contextually similar words and word groups (all the $\phi_k(\delta(i,j))$).

\subsection{Scoring a "parse"}

It was assumed that at each stage of association the most grammatical grouping would generate the greatest number or greatest concentration of grammatically similar elements, and an order of association, branching, or the best phrase structure tree, was selected on that basis. Broadly speaking, for each order of association of words $t$ the greatest of:

\[
max\{\sum_{k} \phi_k(t)\}
\]

\section{Discussion}

The system described in section 2 was the vector parser system as originally formulated between 1998-2001. It was not picked up at the time, although tests were done comparing performance quantitatively for the task of parsing Chinese in 2003. We hope to present these results in a later paper. In general the parser performed on a level comparable with a contemporary symbolic parser, but not greatly better.

It now seems apparent that while classes are not abstracted until runtime in this formulation, important details are lost when word similarity is assessed. Not only class, but similarity between words can, and typically does, vary between contexts. What is needed are vectors of contexts, not similarity based on context.

This should be a fruitful direction for further investigation.

However the method of generating product vectors by an aggressive substitution of contextually similar words and word groups (all the $\phi_k(\delta(i,j))$ above) was quite successful. It enables product vectors to stay in the same space as their operands and makes entire sentences comparable syntactically, and potentially semantically. The vectors generated had sufficient representational strength to generate parse trees at least comparable with contemporary symbolic parsers.

In the context of contemporary explorations of vector combination for semantic representation, this approach of aggressive substitution of contextually similar words and word groups is recommended as a possible solution to problems which present themselves of expansion of representation space, representation strength and comparability.

\begin{acknowledgements}

Many thanks to Wojciech Skut (wojciech@google.com) for the formalism in this paper and more recent comments. His formalization was what made me first realize the "ad-hoc arrangements of examples" model I was proposing could be seen as a kind of vector product.

\end{acknowledgements}

\bibliography{references}

\begin{thebibliography}{1}
  
  \bibitem{chomskyLBLT}  Noam Chomsky, The Logical Basis of Linguistic Theory (1964),	In Proceedings of the Ninth International Congress of Linguists, edited by Horace G. Lunt. The Hague: Mouton \& Co.
  
  \bibitem{chaitinUnknowable} G J Chaitin, The Unknowable, Published by Springer-Verlag Singapore, 1999.

  \bibitem{dagan95contextual} Dagan, Ido, Shaul Marcus and Shaul Markovitch. Contextual word similarity and estimation from sparse data, Computer, Speech and Language, 1995, Vol. 9, pp. 123-152.
  \bibitem{harris-msl} Harris, Z. (1951). Methods in structural linguistics. Chicago, IL: University of Chicago Press).

  \bibitem{two-puzzles}Pawley, A. \& Syder F. (1983) Two puzzles for linguistic theory: nativelike selection and nativelike fluency, in L Richards and IL Schmidt (eds.) 1983: Language and Communication, pp. 191-226. London: Longman.

  \bibitem{powerfultea} Halliday, M.A.K., 1966. Lexis as a linguistic level. In: Bazell, C.E., Catford, J.C., Halliday, M.A.K., Robins, R.H. (Eds.), In Memory of
J.R. Firth, pp. 148–162.
  
  \bibitem{kolmogorovTRN} Kolmogorov, Andrey (1963). "On Tables of Random Numbers". Sankhyā Ser. A. 25: 369–375. MR 178484.
  
  \bibitem{lambAA} Lamb, review of Chomsky ....  American Anthropologist
69.411-415 (1967).

  \bibitem{lambPTP} Lamb, Prolegomena to a theory of phonology.  Language
42.536-573 (1966) (includes analysis of the Russian obstruents
question, as well as a more reasonable critique of the criteria
of classical phonemics).

  \bibitem{lambvanderslice} Lamb and Vanderslice, On thrashing classical phonemics. LACUS
Forum 2.154-163 (1976).

  \bibitem{lambdrummer} Lamb, Linguistics to the beat of a different drummer. First Person Singular III. Benjamins, 1998 (reprinted in Language and Reality, Continuum, 2004).

  \bibitem{lamblinearity} Lamb, Funknet, 24th June, 2004.

  \bibitem{linSim} Dekang Lin, An Information-Theoretic Definition of Similarity, In Proceedings of the 15th International Conference on Machine Learning, 1998, pages 296--304, Morgan Kaufmann.
  
  \bibitem{freemanNAACL} 
  
Freeman R. J., Example-based Complexity--Syntax and Semantics as the 
Production of Ad-hoc Arrangements of Examples, Proceedings of the ANLP/NAACL 
2000 Workshop on Syntactic and Semantic Complexity in Natural Language 
Processing Systems, pp. 47-50. 

  \bibitem{archHilldarling} Archibald A. Hill, "How Many Revolutions Can a Linguist Live Through"; Studies in the History of Linguistics, Vol. 21; 1980

  \bibitem{hopperEG} Paul Hopper, Emergent grammar, Berkeley Linguistics Conference (BLS), 1987, volume 13, pages 139-157. 
 
  \bibitem{lewisLA} Michael Lewis (1993. The lexical approach: The state of ELT and the way forward. Hove, England: Language Teaching Publications.
  
  \bibitem{nattingerLPG} Nattinger, J. (1980), A Lexical Phrase Grammar for ESL. TESOL Quarterly, 14, 337-344.
  
  \bibitem{newmeyerHistPerspective} Frederick J. Newmeyer, Generative Linguistics a historical perspective, Routledge 1996.
  
  \bibitem{weinertRFL} Weinert, R. (1995) The Role of Formulaic Language in Second Language Acquistion: A review. Applied Linguistics 16 (1): 180-205.

  \bibitem{zaanen-bootstrapping} Menno van Zaanen 2001, Bootstrapping Syntax and Recursion using Alignment-Based Learning, Proceedings of the Seventeenth International Conference on Machine Learning. pages 1063-1070

\end{thebibliography}

\end{document}